\documentclass[sigconf,screen]{acmart}
\AtBeginDocument{}

\makeatletter
\def\@ACM@checkaffil{
    \if@ACM@instpresent\else
    \ClassWarningNoLine{\@classname}{No institution present for an affiliation}%
    \fi
    \if@ACM@citypresent\else
    \ClassWarningNoLine{\@classname}{No city present for an affiliation}%
    \fi
    \if@ACM@countrypresent\else
    \ClassWarningNoLine{\@classname}{No country present for an affiliation}%
    \fi
}
\makeatother
    
\usepackage{graphicx}
\usepackage{enumitem}
\usepackage{MnSymbol}
\usepackage{wasysym}
\usepackage{pifont}
\usepackage{multirow}
\usepackage{caption}
\usepackage{subcaption}

\usepackage{xcolor}
\usepackage{algorithmic}
\usepackage[linesnumbered,ruled]{algorithm2e}

\usepackage{graphicx}
\usepackage{makecell}

\iftrue
\usepackage{titlesec}
\titlespacing\section{0pt}{3pt plus 1pt minus 1pt}{0pt plus 1pt minus 1pt}
\titlespacing\subsection{0pt}{3pt plus 1pt minus 1pt}{0pt plus 1pt minus 1pt}
\titlespacing\subsubsection{0pt}{3pt plus 1pt minus 1pt}{2pt plus 1pt minus 1pt}
\fi
\usepackage{enumitem}
\setlist{leftmargin=5.08mm}

\copyrightyear{2024} 
\acmYear{2024} 
\setcopyright{acmlicensed}\acmConference[DAC '24]{61st ACM/IEEE Design Automation Conference}{June 23--27, 2024}{San Francisco, CA, USA}
\acmBooktitle{61st ACM/IEEE Design Automation Conference (DAC '24), June 23--27, 2024, San Francisco, CA, USA}
\acmDOI{10.1145/3649329.3656504}
\acmISBN{979-8-4007-0601-1/24/06}

\begin{document}

\title{GNNavigator: Towards Adaptive Training of Graph Neural Networks via Automatic Guideline Exploration}
\titlenote{
This work is supported in part by National Natural Science Foundation of China (Grant No. 62072019) and National Key Laboratory of Spintronics. 
 Corresponding authors are \textit{Jianlei Yang} and \textit{Chunming Hu}, Email: \url{jianlei@buaa.edu.cn}, \url{hucm@buaa.edu.cn}
}

\author{Tong Qiao}      \affiliation{\institution{Beihang University}}
\author{Jianlei Yang}   \affiliation{\institution{Beihang University}}
\author{Yingjie Qi}     \affiliation{\institution{Beihang University}}
\author{Ao Zhou}        \affiliation{\institution{Beihang University}}
\author{Chen Bai}       \affiliation{\institution{CUHK}}
\author{Bei Yu}         \affiliation{\institution{CUHK}}
\author{Weisheng Zhao}  \affiliation{\institution{Beihang University}}
\author{Chunming Hu}    \affiliation{\institution{Beihang University}}


\begin{abstract}

Graph Neural Networks (GNNs) succeed significantly in many applications recently.
However, balancing GNNs training runtime cost, memory consumption, and attainable accuracy for various applications is non-trivial.
Previous training methodologies suffer from inferior adaptability and lack a unified training optimization solution.
To address the problem, this work proposes GNNavigator, an adaptive GNN training configuration optimization framework.
GNNavigator meets diverse GNN application requirements due to our unified software-hardware co-abstraction, proposed GNNs training performance model, and practical design space exploration solution.
Experimental results show that GNNavigator can achieve up to 3.1$\times$ speedup and 44.9\% peak memory reduction with comparable accuracy to state-of-the-art approaches.

\end{abstract}

\setcopyright{none}

\maketitle
\pagestyle{plain}

\section{Introduction}

Graph neural networks (GNNs) have attained significant success across a wide range of graph-based applications, such as node classification~\cite{kipf2017semi, velivckovic2018graph}, link prediction, community detection and flow forecasting.
Thanks to the information propagation along edges, GNNs exhibit the ability to capture intricate patterns and relationships within graph data, significantly surpassing traditional deep learning approaches.
However, due to neighborhood explosion, GNNs face more serious challenges than traditional deep learning in terms of accuracy, execution time.
Many efforts have been made to address the challenges, which can be generally categorized based on their optimization goals as accuracy centric optimization, time efficiency centric optimization, and memory footprint optimization. 
To minimize feature retrieving traffic, PaGraph~\cite{lin2020pagraph} and BGL~\cite{liu2023bgl} introduce feature caching policies, utilizing free GPU memory for caching.
Works such as FastGCN~\cite{chen2018fastgcn}, GraphSAINT~\cite{zeng2019graphsaint} leverage the locality of graph data for more efficient neighbor sampling~\cite{liu2022gnnsampler}.
To enhance GNNs computation performance, \cite{chen2020fusegnn, wang2021gnnadvisor} develop GPU kernels and thread assignment policies, while~\cite{you2022gcod} design accelerators for GNNs training.
Additionally, there are many other works focused on optimizations such as workflow pipelining~\cite{kaler2022accelerating}, dedicated task scheduling~\cite{yang2022gnnlab}, and feature data compression~\cite{liu2021exact}. 

\begin{figure}[t]
    \centering
    \includegraphics[width=1.0\linewidth]{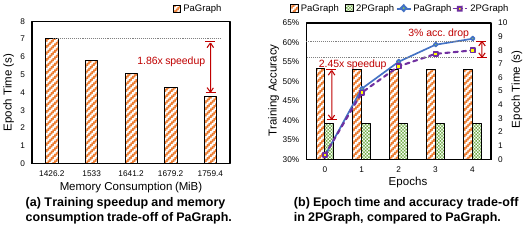}
    \vspace{-16pt}
    \caption{Profiling on existing GNN training frameworks.}
    \label{fig:profile}
    \vspace{-6pt}
\end{figure}

Unfortunately, all aforementioned optimization strategies are not sufficient for tackling existing problems.
Firstly, without a comprehensive view of GNNs training, most existing approaches perform well only under specific scenarios.
As depicted in Fig.~\ref{fig:profile}, existing works typically achieve their claimed excellent performance by making trade-offs among different metrics.
The speedup of PaGraph~\cite{lin2020pagraph} largely depends on the extra consumption of memory resources.
And compared to PaGraph~\cite{lin2020pagraph}, 2PGraph~\cite{zhang20212pgraph} achieves 2.45$\times$ speedup at the cost of a 3\% drop in training accuracy. 
The adaptability of existing works becomes limited when confronted with diverse application requirements and scenario constraints.
Secondly, the collaboration among different optimization strategies is disappointing.
Regardless of the difficulties in strategies combination, simply linking multiple strategies may compromise their performance due to incompatibility.
Finally, many previous works need careful adjustments in configuration to ensure their performance. 
This process largely relies on essential expertise and requires significant human effort.
To this end, enabling automatic exploration for adaptive solutions with low overhead is valuable, according to the varying requirements of graph-based applications. 

In this paper, we introduce a novel GNNs training framework called \textbf{GNNavigator}. 
GNNavigator can automatically generate effective training guidelines based on application requirements. 
Our approach distinguishes itself from other GNNs training optimizations by its adaptability to applications prioritizing different performance metrics.
Furthermore, many existing optimization strategies can be easily reproduced through simple reconfiguration within GNNavigator framework.
Consequently, GNNavigator always achieves excellent performance comparable to or better than previous works.

Our contributions are summarized as follows:
\begin{itemize}
    \item \textbf{Unified optimizations abstraction.} We decompose GNNs training into several components, categorizing and abstracting various optimizations according to the decomposition. 
    \item \textbf{Reconfigurable runtime backend.} Upon the abstractions, we build a reconfigurable runtime backend to support diverse optimizations by simply reconfiguring.
    \item \textbf{``Gray-box'' performance estimating model.} We construct a "gray-box" model, combining theoretical analysis and machine learning, for accurate GNNs training performance estimation.
    \item \textbf{Adaptive training guidelines.} With the assistance of performance estimation, GNNavigator provides training guidelines adaptive to application requirements automatically. 
\end{itemize}

\section{Background and Motivations}

In this section, we outline the problem boundaries of GNNavigator in Sec.~\ref{chapter:minibatch} and Sec.~\ref{chapter:heterogeneous}, and discuss the motivations inspired by several key observations in GNNs training in Sec.~\ref{chapter:observe}.

\subsection{Mini-batch based GNNs Training}
\label{chapter:minibatch}

\begin{algorithm}[t]
\small
    \renewcommand{\algorithmicrequire}{\textbf{Input:}}
    \renewcommand{\algorithmicensure}{\textbf{Output:}}
    \caption{\small Mini-batch based GNNs training on heterogeneous platforms.}
    \label{algo:gnnalgo}
    \begin{algorithmic}[1]
        \REQUIRE graph $G(\mathcal{V}, \mathcal{E})$, batch size $|\mathcal{B}^0|$, initial graph network $M(L, \Phi_{init})$, network layers $L$, network parameters $\Phi$.
        \ENSURE converged graph network $M(L, \Phi_{trained})$.
        
        \STATE \textbf{for} $i$ in [0, $|\mathcal{V}|/|\mathcal{B}^0|)$ \textbf{do}
        \COMMENT {\hspace*{0.2cm}{$\triangleright$ Component 1: Sampling on Host}}
        \STATE \hspace*{0.2cm} $G_i(\mathcal{V}_i, \mathcal{E}_i) \leftarrow \textbf{SubgraphSampling}(G(\mathcal{V}, \mathcal{E}), \mathcal{B}^0_i)$ 
        \COMMENT {\hspace*{0.2cm}{$\triangleright$ Component 2: Transmission}}
        \STATE \hspace*{0.2cm} $\textbf{MemcpyHtoD}(G_i)$
        \COMMENT {\hspace*{0.2cm}{$\triangleright$ Component 3: Computation on Device}}
        \STATE \hspace*{0.2cm} \textbf{for} $l$ in $L$ \textbf{do}
        \STATE \hspace*{0.4cm} $a^l \leftarrow \textbf{Aggregate}(G_i, M)$
        \STATE \hspace*{0.4cm} $h^l \leftarrow \textbf{Combine}(a^l, M)$
        \STATE \hspace*{0.4cm} $loss \leftarrow \textbf{LossFunction}(h^l, G_i)$
        \STATE \hspace*{0.4cm} $\textbf{Backwards}()$
        \STATE \hspace*{0.2cm} \textbf{end for}
        \STATE \textbf{end for}
        \STATE $\textbf{MemcpyDtoH}(M(L, \Phi_{trained}))$
        \STATE \textbf{return} $M(L, \Phi_{trained})$
    \end{algorithmic}
\end{algorithm}

The computation of GNNs can typically be described by their \textit{aggregate} function and \textit{combine} function. 
On a graph $G(\mathcal{V}, \mathcal{E})$ with vertex set $\mathcal{V}$ and edge set $\mathcal{E}$.
$v \in \mathcal{V}$ is a vertex in graph whose feature vector is $h^0_v$, and $\mathcal{N}(v)$ represents its neighborhood.
Let $e^{l-1}_{uv}$ be the edge feature at layer $l-1$.
The computation of a single GNN layer $l$ can be formulated as:
\begin{equation}
\label{eq:gnncompute}
    \begin{split}
        & a_v^l = \textbf{Aggregate}^l \left(h^{l-1}_u, e^{l-1}_{uv} | u \in \mathcal{N}(v) \cup h^{l-1}_v \right) , \\
        & h_v^l = \textbf{Combine}^l(a_v^l) ,
    \end{split}
\end{equation}
where $a_v^l$ is the \textit{aggregate} result and $h_v^l$ is the vertex embedding of $v$ at layer $l$.
By stacking multiple GNN layers together, we can get the final output of graph neural networks~\cite{qi2023architectural}.

To enable GNNs training on tremendously large-scale graphs, mini-batch based training has been introduced.
It conducts training on subgraphs iteratively, to alleviate the ever-increasing requirements in memory.
Mini-batch based training first samples a subgraph $G_i(\mathcal{V}_i, \mathcal{E}_i)$ from $G(\mathcal{V}, \mathcal{E})$ as mini-batch, and then train the network on a series of mini-batches.

\subsection{Heterogeneous Platforms for GNNs Training}
\label{chapter:heterogeneous}

GNN training has been explored across diverse hardware platforms.
The two most influential frameworks, PyG~\cite{Fey/Lenssen/2019} and DGL~\cite{wang2019dgl}, both support GNNs training on heterogeneous architectures like CPU-GPU. 
Aligraph and Euler~\cite{zhu2019aligraph} focus on CPU-only platforms to enable flexible computation patterns.
Many other works use FPGAs~\cite{lin2023hitgnn, lin2022hp} or even accelerators~\cite{you2022gcod, zhou2023hardware} as computation platforms, leading to a notable reduction in time cost or energy consumption.

However, regardless of the diversity in hardware, it is still the mainstream to train GNNs with heterogeneous platforms.
As illustrated in Algo.~\ref{algo:gnnalgo}, complex operations such as sampling and file I/O, are executed on general-proposed platforms such as CPUs, which we call \textit{host}.
Massive but simple operations such as \textit{aggregate} and \textit{combine} are conducted on dedicated designed platforms such as GPUs or FPGAs, which we call \textit{device}.
Furthermore, \textit{host} and \textit{device} can exchange data through \textit{host-device links}, which can be implied through PCIe or DMA. 
Remarkably, based on the assumption that data retrieving within a certain platform is always much faster than fetching data from another platform through data \textit{links}, redundant memory resources on \textit{device} can be treated as a cache to store partial graph data, aiming to accelerate GNNs training~\cite{lin2020pagraph, liu2023bgl}.

\subsection{Observations and Opportunities}
\label{chapter:observe}

Algo.~\ref{algo:gnnalgo} lists the overview of mini-batch based GNNs training on heterogeneous platforms.
The number of mini-batches $|\mathcal{V}|/|\mathcal{B}^0|$ is decided in line $1$.
Given the target vertices set $\mathcal{B}^0_i$ of each iteration, the mini-batch $G_i(\mathcal{V}_i, \mathcal{E}_i)$ is deduced according to the specific sampling algorithm (line $2$). 
The sampled subgraph is then transferred to \textit{device} through \textit{links} between \textit{host} and \textit{device} (line $3$).
Then, GNNs are trained on \textit{device} across the sampled mini-batches (line $4$ to $8$).
The trained model is transferred back to \textit{host} for further processing (line $11$).

We outline $4$ categories of training optimization opportunities based on Algo.~\ref{algo:gnnalgo}, \textit{i.e.},~sampling, transmission, computation, and model design.
Three requirements are summarized, given the limitations of previous related approaches.

\begin{figure*}[t]
    \centering
    \includegraphics[width=0.96\linewidth]{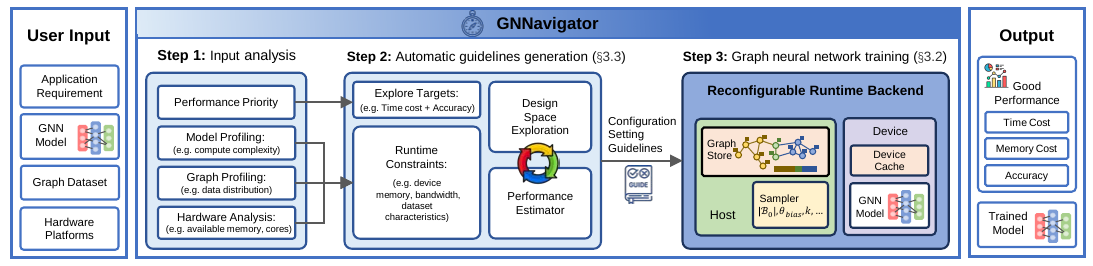}
    \vspace{-12pt}
    \caption{Framework overview of GNNavigator.}
    \label{fig:overview}
    \vspace{-8pt}
\end{figure*}

\textbf{Compatibility.}
The framework should be compatible with many dedicatedly designed GNN training optimizations.
For example, FPGA-orient optimization~\cite{lin2022hp} is orthogonal to GNNAdvisor~\cite{wang2021gnnadvisor}.
However, compatibility permits a feasible joint optimization by combining these two methodologies.

\textbf{Adaptability.}
The framework should be adaptable to various applications.
Different GNN applications pertain to various characteristics, emphasizing runtime performance or hardware budgets.
It is non-trivial to find a sweet one-for-all solution.
Adaptability allows the framework to produce optimal training optimization strategies given different scenarios.

\textbf{Automation.}
The framework should be automated. The automation alleviates heavy labor force input in deciding optimal training optimization parameters.
Inspired by BOOM-Explorer~\cite{bai2021boom}, we formulate the automation process as a design space exploration (DSE) problem and solve it via our customized surrogate model.

\section{GNNavigator Framework}

Motivated by the observations and opportunities outlined in Sec.~\ref{chapter:observe}, we introduce \textbf{GNNavigator}, an adaptive framework automatically fine-tuning GNNs training according to application requirements and hardware constraints.
GNNavigator is constructed upon three pivotal techniques: 1) a unified and reconfigurable backend facilitating efficient strategy cooperation, 2) a "gray-box" performance estimator, and 3) an application-driven design space exploration tailored to requirements and constraints.

\subsection{Framework Overview}

Fig.~\ref{fig:overview} provides an overview of GNNavigator, and its general workflow.
For better adaptability, GNNavigator requires some essential information, typically related to the applications, as input.

Users should specify the following input items:
\begin{itemize}
    \item The graph dataset $G(\mathcal{V}, \mathcal{E})$ to be trained on.
    \item The GNN model $M(L, \Phi_{init})$, with explicit network architecture.
    \item The application requirements like time cost $T$, memory consumption $\Gamma$, accuracy $Acc$, etc., along with the user-defined priorities for different requirements.
    \item The heterogeneous hardware platforms for GNNs training.
\end{itemize}

The user inputs are quantitatively analyzed to formulate the parameterized \textit{explore targets} and \textit{runtime constraints}, as shown in Step 1.
Then, in Step 2, GNNavigator automatically generates GNN training guidelines, taking both \textit{explore targets} and \textit{runtime constraints} into account to ensure its adaptability.
We further design a "gray-box" performance estimator to accurately predict the training performance with relatively low overhead. 
Users receive the guidelines, in the form of training configuration settings, and apply these settings on GNNavigator's runtime backend for GNNs training (Step 3).
GNNavigator guarantees that the actual training performance $Perf\{T, \Gamma, Acc\}$, measured in terms of time cost $T$, memory consumption $\Gamma$, and accuracy $Acc$, not only satisfies application requirements but also outperforms other handcrafted designs.

\subsection{\mbox{Unified Abstraction of Training Optimizations}}
\label{chapter:backend}

\begin{figure}[t]
    \centering
    \includegraphics[width=0.96\linewidth]{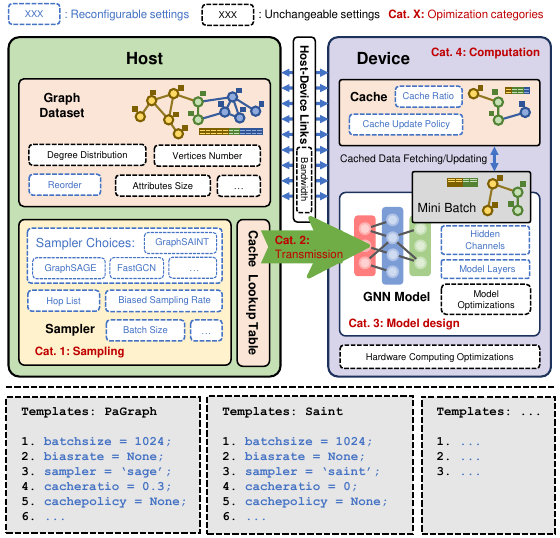}
    \vspace{-8pt}
    \caption{Reconfigurable runtime backend of GNNavigator.}
    \label{fig:backend}
    \vspace{-8pt}
\end{figure}

The various GNNs training optimizations can be generally classified into four categories: \textbf{sampling strategies}, \textbf{transmission strategies}, \textbf{computation optimizations}, and \textbf{model design optimizations}, according to the decomposition of Algo.~\ref{algo:gnnalgo}, as introduced in Sec.~\ref{chapter:observe}.
Unified abstractions for optimizations in each category are generated respectively.
Furthermore, as shown in Fig.~\ref{fig:backend}, a reconfigurable runtime backend is established based on the categorizations and abstractions, with optimizations from different categories mapping to different parts of the backend.

\textbf{Sampling strategies.} 
There are node-wise samplers, layer-wise samplers, and subgraph-wise samplers for unbiased sampling~\cite{liu2022gnnsampler}, and dedicated designed locality-aware samplers aiming for biased sampling~\cite{zhang20212pgraph}.
Despite the diversity in sampling strategies, samplers generally expand a subgraph from given target vertex set.
Therefore, we can provide a unified abstraction of sampling strategies, that is, samplers iteratively fanout vertices at certain probability, and further generate subgraphs. 

In our abstraction of samplers, it receives a certain number of target vertices $\mathcal{B}_i^{l-1}$ as input from layer $l-1$, and fanouts $k^l$ neighbors from every vertex $v_i^{l-1} \in \mathcal{B}_i^{l-1}$ at a given probability. 
The output $\mathcal{B}_i^{l}$ at layer $l$ can be formulated as follows:
\begin{equation}
    \mathcal{B}^l_i = \bigcup_{v_i^{l-1} \in \mathcal{B}^{l-1}_i} u \cdot \mathbb{I}_{p(\eta)}\left(\frac{k^l}{|\mathcal{N}(v_i^{l-1})|}\right) , u\in \mathcal{N}(v_i^{l-1}).
    \label{eq:sample}
\end{equation}
$\mathbb{I}_{p(\eta)}$ is an indicator function to decide whether to select a neighbor $u$, according to a probability $p(\eta)$ specified by the sampling algorithm.

While Eq.~\ref{eq:sample} is primarily in the form of node-wise sampling, it can generalize other sampling strategies as well.
For instance, in the case of layer-wise sampling~\cite{chen2018fastgcn}, the number of sampled nodes at layer $l$ can be represented as $\Delta^l$, which is a predetermined value.
We can derive the mathematical expectation of $k^l$ from $\Delta^l$ by:
\begin{equation}
\label{eq:ekl}
    \mathbb{E}(k^l)=\frac{\Delta^l}{|\mathcal{B}_i^{l-1}|} \cdot \mu\left(p(\eta), \mathcal{B}_i^{l-1}\right) ,
\end{equation}
where $\mu(p(\eta), \mathcal{B}_i^{l-1})$ is a coefficient which indicates the probability of multi-vertices in $\mathcal{B}_i^{l-1}$ shares a common neighbor in $\mathcal{B}_i^l$.
In this way, layer-wise sampling has been uniformly abstract as Eq.~\ref{eq:sample}.
Locality-aware sampling and subgraph-wise sampling can be more easily integrated into the abstraction, according to their sampling patterns.
By setting the neighbor selection probability to a function of data locality $p(\eta)$, we can reproduce biased samplers that prefer a certain subset of $\mathcal{V}$.
Subgraph-wise sampling strategies like GraphSAINT~\cite{zeng2019graphsaint} can be viewed as a special case of node-wise sampling, with many more hops, but only a single neighbor fanout in each hop.
To this end, we can unify different sampling strategies to the sampler in Fig.~\ref{fig:backend}, with configurable settings being enumerated  

\textbf{Transmission strategies.} 
Regardless of the implementation of transmission strategies, they always ensure the required data being on the \textit{device} when computing. 
Note that not all required data needs transmission.
An abstraction can be drawn, according to the gap between the required data volume and actual transferred data volume.
The transmission strategies typically leverage the free memory resources on \textit{device} as a cache to alleviate redundant data transmission.
Despite their substantial differences in cache updating policies, 
we can consistently abstract them as follows.
First, the device cache is initialized according to the available memory resource on \textit{device}.
Given a mini-batch, the device cache figures out which part of the mini-batch has been cached.
The remaining part is filtered out from the \textit{host} and transferred to the \textit{device} through \textit{host-device links}.
With all essential data on \textit{device}, training on the mini-batch can be conducted.
Finally, the device cache is updated according to the cache updating policy.
Uniformly, part of the configurable settings on transmission are listed in the device cache in Fig.~\ref{fig:backend}.
We can distinguish different transmission strategies by configuring the settings properly.

\textbf{Computation and model design optimizations.}
GNN models typically embed graph topological information through \textit{aggregate} functions and enable feature learning by \textit{combine} functions.
Let along the detailed design of models,  abstraction of model design optimizations can be formulated based on the time complexity and spatial complexity of \textit{aggregate} and \textit{combine} functions, as shown in Eq.~\ref{eq:gnncompute}.
Similarly, computation optimizations are abstracted by their maximum throughput and available memory resources.
Although the computation optimizations show significant diversity in their targeting \textit{device} platforms, ranging from GPUs to FPGAs, we find that they can all be measured by their computing capability and available resources. 

In Fig.~\ref{fig:backend}, both computation optimizations and model design optimizations are mapped to the \textit{device}, for actual execution and the following performance estimation.

\textbf{Reconfigurable runtime backend.}
Thanks to its reconfigurability, the runtime backend can represent itself with diverse performance, making it adaptive to a wide range of applications.

Fig.~\ref{fig:backend} depicts the mapping relationships between the four optimization categories and backend components, marked by red words. 
Within each backend component, there are many reconfigurable settings that can be freely adjusted to simulate existing approaches, marked with blue dash-line rectangles.
For instance, by disabling \textit{cache update policy}, and properly configuring the \textit{cache ratio}, the backend generally reproduces the approach proposed in PaGraph~\cite{lin2020pagraph}.
Similarly, many existing works can be conveniently reproduced, by applying the configurations setting templates shown in Fig.~\ref{fig:backend}
Furthermore, the unified runtime backend allows for a flexible combination of optimizations from different categories and greatly outperforms existing works in its compatibility.
Note that the runtime backend can even incrementally support future optimizations only if they submit to our abstraction. 
Additionally, all reconfigurable parameters in the runtime backend make up the design space, which will be discussed in detail in Sec.~\ref{chapter:auto}.

\subsection{Automatic Guidelines Generation}
\label{chapter:auto}

\begin{figure}[t]
    \centering
    \includegraphics[width=0.95\linewidth]{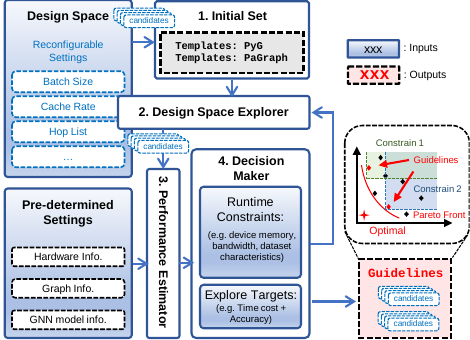}
    \vspace{-8pt}
    \caption{Automatic guidelines exploration.}
    \label{fig:automatic}
    \vspace{-6pt}
\end{figure}

GNNavigator leverages multi-objective design space exploration (DSE) to automatically generate the training guidelines, as shown in Fig.~\ref{fig:automatic}. 
It benefits from multi-objective DSE in two key aspects.
Firstly, in terms of fine-tuning GNN training,  the substantial burden of human labor is mitigated through automatic exploration.
Secondly, in terms of adaptability, the explorer generates guidelines that satisfy user demands by emphasizing different \textit{explore targets}.
Notably, all the reconfigurable settings in Fig.~\ref{fig:backend} constitute the design space, represented by blue dash-line rectangles.
Moreover, to accelerate the exploration, a "gray-box" performance estimator is established based on the unified abstractions of optimization strategies.

\textbf{Gray-box performance estimator.}
The estimator predicts GNN training performance in a "gray-box" manner, combining purely theoretical analysis (white-box) and machine learning methods (black-box) together.

As shown in Fig.~\ref{fig:automatic}, the estimator makes predictions based on 1) the specific values of all configurable settings, which will be represented as \textit{candidate} in design space in our following illustration and 2) the pre-determined settings in runtime, usually determined by applications.
To ensure the accuracy of estimation, the estimator first theoretically analyzes the data dependence between its inputs and performance $Perf\{T, \Gamma, Acc\}$. 
Considering the complexity and randomness of graph, black-box models based on machine learning are introduced to estimate some key intermediate variables that influence $T$, $\Gamma$, and $Acc$.
We will present the methodology of constructing "gray-box" models on performance $Perf$ as follows.

Note that the operations on \textit{device} are independent of those on \textit{host}.
The epoch time can be formulated as:
\begin{equation}
    T = n_{iter} \cdot max\left(t_{sample}+t_{transfer},t_{replace}+t_{compute}\right) ,
\end{equation}
where $t_{sample}$, $t_{transfer}$, $t_{replace}$, $t_{compute}$ represent the time cost of sampling, transmission, cache updating, and computation on \textit{device}, respectively, and $n_{iter}$ is the number of mini-batches within an epoch. 
Let us begin with $t_{replace}$. In scenarios requiring cache replacement, the cache updating overhead is mainly influenced by cache volume $r \cdot |\mathcal{V}|$, and volume of replaced stale data $|\mathcal{V}_i|(1-hit)$.
\begin{equation}
    t_{replace} = f_{replace}\left(r|\mathcal{V}|, |\mathcal{V}_i|(1-hit), Device\right) .
    \label{eq:t_rep}
\end{equation} 
The $hit$ represents the average cache hit rate, and we use $Host$ $Device$ to indicate the hardware information of \textit{host} and \textit{device} respectively.

In a similar fashion, $t_{transfer}$ is determined by the volume of data awaiting transmission $n_{attr}|\mathcal{V}_i|(1-hit)$.
$t_{sample}$ is primarily affected by changes in subgraph size $|\mathcal{V}_i|-|\mathcal{B}^0|$, which represents the transition from the original target vertices to the final mini-batch.
Lastly, mini-batch size $|\mathcal{V}_i|$ and the GNN model $M(L, \Phi)$ together determine $t_{compute}$.
We demonstrate the formulations of $t_{replace}$, $t_{transfer}$, and $t_{sample}$ as follows:
\begin{equation}
    t_{transfer} = f_{transfer}(n_{attr}|\mathcal{V}_i|(1-hit), Host, Device) ,
\end{equation}
\begin{equation}
    t_{sample} = f_{sample}(|\mathcal{V}_i|-|\mathcal{B}^0|, Host) ,
\end{equation}
\begin{equation}
    t_{compute} = f_{compute}(\mathcal{V}_i, M, Device) .
\end{equation}
The term $n_{attr}$ denotes the attribute dimensions of an individual node.

Remarkably, the functions $f_{compute}$, $f_{replace}$, $f_{transfer}$, and $f_{sample}$ can all be estimated using a pre-trained black-box model. 
In this way, the performance estimator can predict the execution time of GNN training with negligible latency.

The prediction of device memory consumption $\Gamma$, can also be decomposed to sub-tasks of estimating $\Gamma_{model}$, $\Gamma_{cache}$, $\Gamma_{runtime}$ respectively,
\begin{equation}
    \Gamma = \Gamma_{model} + \Gamma_{cache} + \Gamma_{runtime} ,
    \label{eq:mem_decom}
\end{equation}
where $\Gamma_{model}$, $\Gamma_{cache}$, $\Gamma_{runtime}$ are formulated as follows:
\begin{equation}\begin{split}
    & \Gamma_{model} \propto |\Phi| , \\
    & \Gamma_{cache} = f_{cache}(r|\mathcal{V}|n_{attr}) , \\
    & \Gamma_{runtime} = f_{runtime}(\overline{|\mathcal{V}_i|}, \Phi) .
\end{split}\end{equation}
$\Gamma_{model}$ reveals the static memory consumption of GNNs, directly related to $|\Phi_{init}|$. 
$\Gamma_{cache}$ represents the cache memory consumption, and $\Gamma_{runtime}$ indicates the memory footprint of mini-batch computation phrase.

Estimation of model accuracy falls back behind the ones on the other two metrics in its explainability.
Nevertheless, we try to analyze it from the perspective of data distribution.
Taking the training accuracy on mini-batches with unbiased sampling as the baseline, the estimator measures the accuracy changes $\delta_{Acc}$ of training as:
\begin{equation}
    \delta_{Acc} = f_{accuracy}(Deg(G_i), Deg(G), |\mathcal{V}_i|) .
    \label{eq:accu}
\end{equation}
The formulation on accuracy changes is established upon the assumption that a mini-batch will learn more information about a given graph $G$ by focusing on the vertices with more importance.
However, the prediction on accuracy is still more like a black box, compared with $T$ and $\Gamma$.

\begin{figure}[t]
    \centering
    \includegraphics[width = 1.0\linewidth]{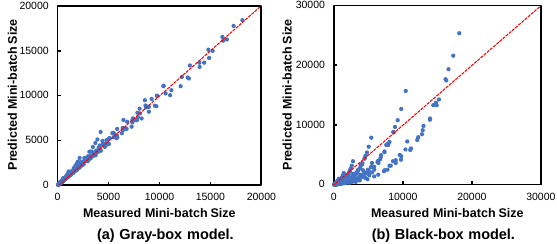}
    \vspace{-16pt}
    \caption{Accuracy comparison between different estimator models. The distance between each blue point and the red dash-line reflects the estimator’s accuracy}
    \label{fig:estimation}
    \vspace{-8pt}
\end{figure}

Notably, as can be witnessed in the theoretical analysis, the mini-batch size $|\mathcal{V}_i|$ plays an important role in performance estimation.
Considering its significance, we analytically formulate the expectation of mini-batch size $\mathbb{E}(|\mathcal{V}_i|)$ as: 
\begin{equation}
    \label{eq:mini-batch}
    \mathbb{E}\left(|\mathcal{V}_i|\right) = 
    f_{overlapping}\left(|\mathcal{B}^0|\prod^L_{l=1}{(1+k^l)^\tau}, p(\eta)\right) .
\end{equation}
where $f_{overlapping}$ is a penalty function determined by graph characteristics and is certainly learnable.
Consequently, we obtain an extremely accurate prediction of mini-batch size by Eq.~\ref{eq:mini-batch}, far better than the pure black-box model (Decision Tree Regression), as shown in Fig.~\ref{fig:estimation}. The predicted values and the measured values are more consistent with the equal line, which is marked with red dash-line.

\textbf{Application-driven design space exploration.}
Based on GNNavigator's precise estimation of performance, we can conduct the design space exploration that is adaptive to hardware constraints and application requirements. 

As shown in Fig.~\ref{fig:automatic}, the explorer starts its exploration from an initial \textit{candidates} configured according to templates of existing works, to ensure the generated guidelines could achieve at least a comparable performance to previous approaches.
Then, it travels across all configurable settings, with the depth-first-search (DFS) algorithm. 
The explorer iteratively tries \textit{candidates} in design space and gets the performance of \textit{candidates} through the performance estimator. 
According to the \textit{explore targets}, the \textit{decision maker} determines whether to accept a \textit{candidate} as the guideline or to continue the exploration.
Notably, some \textit{runtime constraints} are imposed according to different applications.
The explorer will prune the design space to accelerate the process of exploration when the estimated performance cannot satisfy the \textit{runtime constraints}.
With an awareness of application requirements, the explorer emphasizes the specific performance metrics and leverages Pareto front theory to obtain the most suitable \textit{candidates}, which are finally provided as training guidelines.

\section{Experimental Results}

\subsection{Evaluation Setup}

\textbf{Baselines.} To evaluate the effectiveness of GNNavigator, we consider 3 baselines:
PyG~\cite{Fey/Lenssen/2019}, a state-of-the-art GNN framework on both CPUs and GPUs;
PaGraph~\cite{lin2020pagraph}, a GNNs computation framework with static cache; 
2PGraph~\cite{zhang20212pgraph}, a CPU-GPU heterogeneous framework, with cache-aware sampling to accelerate training.
PaGraph~\cite{lin2020pagraph}, 2PGraph~\cite{zhang20212pgraph}, and original PyG~\cite{Fey/Lenssen/2019} are all reproduced on our runtime backend for a fair comparison.
The reproductions achieve similar results as the ones they report.
Note that the volume of available GPU memory will significantly influence the performance of PaGraph~\cite{lin2020pagraph}.
Therefore, we measure the performance of PaGraph~\cite{lin2020pagraph} under ideal circumstances (Pa-Full) and resource-limited circumstances (Pa-Low) respectively.

\textbf{Datasets and platforms.} Our experiments are conducted on datasets with various scales, including Ogbn-arxiv (AR), Ogbn-products (PR)~\cite{hu2020ogb}, Reddit2 (RD2), and representative graph neural networks, including GCN, GAT, GraphSAGE (SAGE).
We test the performance of the runtime backend on different devices such as RTX 4090, A100, and M90, and further set manual constraints to simulate various scenarios of application.
The time cost and memory footprint are measured by  PyTorch profiler~\cite{paszke2019pytorch}.

\textbf{Performance estimator settings.} 
The performance estimator is trained on the ground-truth performance covering the whole design space.
For fairness, the estimator is established upon the performance across all the datasets available, except the one waiting for estimation. 
Specifically, to embed more prior knowledge, we randomly generate some power-law graphs and profile the training on them as data enhancement to optimize our performance estimator.

\begin{table}[t]
    \centering
    \caption{Performance of GNNavigator across different tasks.}
    \vspace{-6pt}
    \resizebox{0.48\textwidth}{!}{
    \begin{tabular}{c|c|c|c|c}
    \hline
        \makecell[c]{Applications \\
        (Dataset+Model)}  & Method & Time ($T$)/s & \makecell[c]{Memory ($\Gamma$)/GB} & Accuracy ($Acc$) \\ \hline
        \multirow{8}{*}{PR + SAGE} & PyG~\cite{Fey/Lenssen/2019} & 9.27 & 1.25 & 90.55\% \\
                             & Pa-Full~\cite{lin2020pagraph} & 5.44(1.7$\times\uparrow$) & 2.11(69.1$\%\uparrow$) & 90.42\% \\
                              & Pa-low~\cite{lin2020pagraph} & 8.39(1.1$\times\uparrow$) & 1.34(7.5$\%\uparrow$) & 90.58\% \\ 
                                  & 2P~\cite{zhang20212pgraph} & 5.18(1.8$\times\uparrow$) & 0.88(29.7$\%\downarrow$) & 90.36\% \\ 
                                  & \textbf{Bal} & \textbf{3.67(2.5$\times\uparrow$)} & \textbf{1.23(7.5$\%\downarrow$)} & \textbf{91.19\%} \\ 
                               & \textbf{Ex-TM} & \textbf{3.95(2.3$\times\uparrow$)} &  \textbf{0.78(37.3$\%\downarrow$)} & 90.37\% \\
                               & \textbf{Ex-MA} & 5.12(1.8$\times\uparrow$) &  \textbf{0.88(29.7$\%\downarrow$)} & \textbf{91.22\%} \\
                               & \textbf{Ex-TA} & \textbf{3.59(2.6$\times\uparrow$)} & 1.64(31.1$\%\uparrow$) & \textbf{91.24\%} \\ \hline
        \multirow{8}{*}{RD2 + SAGE} & PyG~\cite{Fey/Lenssen/2019} & 7.68 & 1.32 & 79.28\% \\
                             & Pa-Full~\cite{lin2020pagraph} & 3.78(2.0$\times\uparrow$) & 1.80(36.3$\%\uparrow$) & 79.23\% \\
                              & Pa-low~\cite{lin2020pagraph} & 7.04(1.1$\times\uparrow$) & 1.43(7.6$\%\uparrow$) & 79.25\% \\ 
                                  & 2P~\cite{zhang20212pgraph} & 3.51(2.2$\times\uparrow$) & 0.84(36.36$\%\downarrow$) & 75.95\% \\ 
                                  & \textbf{Bal} & \textbf{3.53(2.1$\times\uparrow$)} & \textbf{0.87(34.1$\%\downarrow$)} & \textbf{80.03\%} \\ 
                               & \textbf{Ex-TM} & \textbf{2.45(3.1$\times\uparrow$)} & \textbf{0.98(44.9$\%\downarrow$)} & 76.42\% \\
                               & \textbf{Ex-MA} & 3.82(2.0$\times\uparrow$) & \textbf{0.91(31.1$\%\downarrow$)} & \textbf{81.16\%} \\
                               & \textbf{Ex-TA} & \textbf{2.85(2.7$\times\uparrow$)} & 0.99(25.0$\%\downarrow$) & \textbf{79.87\%} \\ \hline
        \multirow{8}{*}{AR + GAT} & PyG~\cite{Fey/Lenssen/2019} & 3.49 & 5.80 & 61.44\% \\
                             & Pa-Full~\cite{lin2020pagraph} & 2.98(1.2$\times\uparrow$) & 5.87(1.3$\%\uparrow$) & 61.38\% \\
                              & Pa-low~\cite{lin2020pagraph} & 3.46(1.0$\times\uparrow$) & 5.80(0.1$\%\uparrow$) & 61.45\%\\ 
                                  & 2P~\cite{zhang20212pgraph} & 3.53(1.0$\times\uparrow$) & 5.81(0.2$\%\uparrow$)  & 60.51\% \\ 
                                  & \textbf{Bal} & \textbf{2.98(1.2$\times\uparrow$)} & \textbf{5.87(1.3$\%\uparrow$)} & \textbf{61.43\%} \\ 
                               & \textbf{Ex-TM} & \textbf{3.21(1.1$\times\uparrow$)} & \textbf{5.84(0.8$\%\uparrow$)} & 61.07\% \\
                               & \textbf{Ex-MA} & 3.23(1.1$\times\uparrow$) & \textbf{5.85(0.9$\%\uparrow$)} & \textbf{61.71\%} \\
                               & \textbf{Ex-TA} & \textbf{2.98(1.2$\times\uparrow$)} & 5.87(1.3$\%\uparrow$) & \textbf{61.43\%} \\ 
    \hline
    \end{tabular}
    }
    \label{tab:overall}
    \vspace{-6pt}
\end{table}

\subsection{Overall Performance}

GNNavigator provides guidelines from a comprehensive view of performance $Perf\{T, \Gamma, Acc\}$. 

As shown in Tab.~\ref{tab:overall}, when highlighting a balance among $T, \Gamma, Acc$, GNNavigator generally achieves similar or superior performance compared to the baselines across various GNNs training tasks, denoted as "balance" (\textbf{Bal}).
GNNavigator can further improve training performance in certain metrics, with a marginal trade-off in others, which is marked as "extreme" (\textbf{Ex}) in Tab.~\ref{tab:overall}.
And, \textbf{Ex-TM}, \textbf{Ex-MA}, \textbf{Ex-TA} denote the different priorities of generated guidelines.
For example, \textbf{Ex-TM} emphasizes time $T$ and memory $\Gamma$, leading to up to 3.1$\times$ sppedup and 44.9\% reduction in $\Gamma$, with a negligible drop in $Acc$ by 2.8\%.
On average, GNNavigator achieves 2.3$\times$ acceleration, 27\% reduction in $\Gamma$, across various GNN training tasks. 
Additionally, it also outperforms many other state-of-the-art works~\cite{wang2021gnnadvisor, liu2023bgl}, according to the results they report.

Overall, GNNavigator consistently achieves excellent performance, adaptive to diverse performance priorities.
It outperforms PyG, PaGraph, and 2PGraph in acceleration by up to 3.1$\times$, 2.8$\times$, and 1.4$\times$ respectively.
GNNavigator achieves a significant reduction in memory cost by up to 44.9\% when compared with state-of-the-art works. 
Note that PaGraph~\cite{lin2020pagraph} will bring in extra memory overhead.

\begin{figure}[t]
    \centering
    \includegraphics[width = 1.0\linewidth]{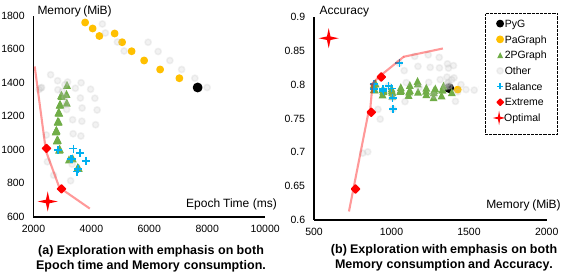}
    \vspace{-18pt}
    \caption{Adaptability validation of generated guidelines on Reddit2+SAGE.}
    \label{fig:trade-off}
    \vspace{-3pt}
\end{figure}

\subsection{Impact of Application Adaption}

The trade-offs among three performance metrics are reported in Fig.~\ref{fig:trade-off}.
The performance statistics are collected by actually executing the training under different configuration settings from the design space.
Design space has been exhausted and each point in Fig.~\ref{fig:trade-off} denotes the performance of a certain candidate in design space.
Furthermore, we manually draw the Pareto front in red lines.
The ground-truth performance of \textit{extreme} (\textbf{Ex}) is marked with red color, and the ones of \textit{balance} (\textbf{Bal}) are marked with blue color.
The adaptability of GNNavigator can be validated since the provided guidelines can perfectly match the actual Pareto front.

Notably, the GNNavigator always takes approaches of existing works into consideration.
Therefore, it will certainly recommend a reproduction of one existing approach as a guideline, if the approach just outperforms others under a given scenario.

\subsection{Precision of Performance Estimator}

Different evaluation metrics including \textit{R2 Score} and \textit{Mean Square Error (MSE)}, are adopted to measure the precision of estimations in $T$, $\Gamma$, and $Acc$, according to their different methods of predicting.
It is convinced that \textit{R2 Score} is more suitable for models with relatively clear theoretical analysis, and \textit{MSE} fits black-box models better.
The results in Tab.~\ref{tab:correctness} show that the "gray-box" estimator can precisely foretell training performance in a low-latency way across a wide range of datasets. 
Bear in mind that \textit{R2 Scores} indicate better precision of estimators when they are closer to 1.
The \textit{R2 Scores} of $T$ and $\Gamma$ range from 0.72 to 0.98, in terms of estimation on Reddit, Reddit2, and Ogbn-products.
And \textit{MSE} of $Acc$ estimation are controlled to a relatively low level, that is 0.03 in the worst case.
These results strongly validate the effectiveness and correctness of our performance estimator.

\begin{table}[t]
    \centering
    \caption{Validation of estimator prediction.}
    \vspace{-8pt}
    \resizebox{0.48\textwidth}{!}{
    \begin{tabular}{c|c|c|c|c}
    \hline
        \makecell[c]{Prediction \\ Validation} & Performance Metric & Reddit & Reddit2 & Ogbn-products \\ \hline
        \multirow{2}{*}{\textit{R2 Score}} & Time Cost ($T$) & 0.8371 & 0.7328  & 0.7281          \\ \cline{2-5}
         & \makecell[c]{Memory ($\Gamma$)} & 0.9240 & 0.9810 & 0.7307   \\ \hline
        \textit{MSE} & Accuracy ($Acc$) &  0.0292 & 0.0249 & 0.0156  \\
        \hline
    \end{tabular}
    }
    \label{tab:correctness}
    \vspace{-6pt}
\end{table}

\section{Conclusions}

We present GNNavigator, a GNN training framework with excellent adaptability to various application requirements.
GNNavigator automatically generates training guidelines, consistently delivering promising performance across different scenarios.
We draw unified abstractions from various optimizations and build a reconfigurable runtime backend based on the abstractions.
To accurately predict training performance on our backend, we construct a gray-box performance estimator.
GNNavigator further enables automatic exploration of training guidelines adapted to application requirements.
Our experiments demonstrate that GNNavigator outperforms state-of-the-art works by up to 3.1$\times$ speedup and at most 44.9\% reduction in memory consumption, with comparable accuracy.

\bibliographystyle{unsrt}
\bibliography{ref}

\end{document}